\documentclass{article}

\PassOptionsToPackage{numbers, compress}{natbib}
\usepackage[preprint]{neurips_2026}

\usepackage[utf8]{inputenc} 
\usepackage[T1]{fontenc}    
\usepackage{hyperref}       
\usepackage{url}            
\usepackage{booktabs}       
\usepackage{enumitem}       
\usepackage{amsfonts}       
\usepackage{nicefrac}       
\usepackage{microtype}      
\usepackage{xcolor}         
\usepackage{graphicx}       
\usepackage{wrapfig}        
\usepackage[most]{tcolorbox} 
\usepackage{multirow}
\usepackage{booktabs}
\usepackage[table]{xcolor}
\usepackage{colortbl}
\usepackage{tablefootnote}

\title{LiteCoder-Terminal: Scaling Long-Horizon Terminal Environments for Learning Language Agents}

\author{
  Xiaoxuan Peng${}^{1,2*}$ \And
  \textbf{Kaiqi Zhang}${}^{1,2*}$ \And
  \textbf{Xinyu Lu}${}^{1,2*}$ \AND
  \textbf{Boxi Cao}${}^{1}$ \And
  \textbf{Yaojie Lu}${}^{1}$ \And
  \textbf{Hongyu Lin}${}^{1}$ \And
  \textbf{Xianpei Han}${}^{1}$ \And
  \textbf{Le Sun}${}^{1}$ \AND
  \normalfont${}^{1}$Chinese Information Processing Laboratory,\\ Institute of Software, Chinese Academy of Sciences \\
  ${}^{2}$University of Chinese Academy of Sciences \\
  {\tt \{pengxiaoxuan2026,zhangkaiqi2024,luxinyu2021\}@iscas.ac.cn}
}

\begin{document}

\maketitle
\begingroup
\renewcommand\thefootnote{*}
\footnotetext{Equal contribution.}
\endgroup

\begin{center}
\vspace{-2.5em}
\raisebox{-2pt}{\includegraphics[height=1.2em]{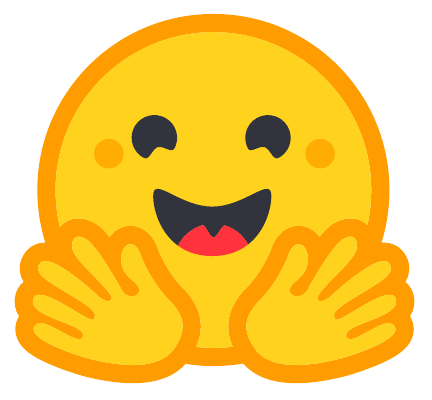}}\,\textbf{Models \& Datasets:}\quad\url{https://huggingface.co/Lite-Coder/}
\vspace{0.5em}
\end{center}

\begin{abstract}
Mastering terminal environments requires language agents capable of multi-step planning, feedback-grounded execution, and dynamic state adaptation. However, training such agents is currently bottlenecked by a reliance on scraped external repositories, which limits domain diversity, environment controllability, and the targeting of specific capability deficits. We introduce \texttt{LiteCoder-Terminal-Gen}, a zero-dependency synthesis pipeline that autonomously generates executable and verifiable terminal training environments directly from domain specifications. Using this framework, we construct two large-scale resources: \texttt{LiteCoder-Terminal-SFT}, comprising 11,255 expert trajectories across 10 domains, and \texttt{LiteCoder-Terminal-RL}, featuring 602 verifiable environments for trajectory-level preference optimization. Supervised fine-tuning of Qwen-family models on our SFT dataset yields agents that significantly outperform their base counterparts. Notably, our 32B variant achieves 29.06\%, 18.54\%, and 34.00\% pass@1 on Terminal Bench 1.0, 2.0, and Pro, respectively. Furthermore, applying Direct Multi-turn Preference Optimization (DMPO) on our RL environments yields additional performance gains. These results systematically demonstrate that fully synthetic, executable environments offer a scalable and verifiable supervision signal for mastering complex, real-world command-line workflows.
\end{abstract}

\section{Introduction}
Recent advancements~\citep{yao2022react, schick2023toolformer, claude_code_2025} have empowered Large Language Models (LLMs) to transition from conversational assistants~\citep{ouyang2022training, touvron2023llama} into autonomous agents capable of interacting dynamically with complex digital environments~\citep{zhou2023webarena, xie2024osworld, feng2026longcli}. Among these environments, the Command-Line Interface (CLI) represents the most general-purpose and foundational interface for digital interaction.

Driven by this shift, the community urgently requires scalable methods to generate diverse terminal environments for both learning and evaluating. Unlike the patch generation tasks evaluated in SWE-bench~\citep{yang2025swe}, terminal-based tasks—as pioneered by Terminal Bench~\citep{merrill2026terminal}—situate agents in a partially observable environment that necessitates a robust capacity to manage complex system changes, demanding both dynamic environment adaptation and persistent goal orientation across long-horizon interactions.

In response to this urgent demand, we introduce LiteCoder-Terminal-Gen, a zero-dependency terminal environment synthesis framework. LiteCoder-Terminal-Gen features an end-to-end pipeline to construct diverse terminal environments and expert demonstrations entirely from scratch. Specifically, the synthesis process operates through three core stages: (1) given a target skill definition detailing an area where the model requires improvement, the framework autonomously generates a massive scale of expert-level task drafts; (2) from these vast propositions, it dynamically instantiates the appropriate underlying terminal environments required for task execution; and (3) grounded in these established tasks and environments, it automatically constructs robust test cases to provide fine-grained scoring criteria.

Crucially, this zero-dependency architecture represents a fundamental departure from existing synthesis pipelines. It eliminates the labor-intensive process of scraping, filtering, and curating high-quality issues from massive external sources like GitHub or Stack Overflow. By breaking free from the constraints of human-curated data repositories, LiteCoder-Terminal-Gen enables a highly targeted training paradigm: it can actively generate specific training environments and trajectories on-demand to directly address and overcome an agent's identified capability deficits.

Starting from these synthesized tasks, we build \texttt{LiteCoder-Terminal-SFT}, a collection of 11,255 expert trajectories generated with capable teacher models like MiniMax models, and fine-tune three Qwen-family base models from 4B to 32B scales. The resulting \texttt{LiteCoder-Terminal} models demonstrate strong proficiency in complex, long-horizon system operations across model scales. In particular, our best-performing 32B model achieves 29.06\%, 18.54\%, and 34.00\% pass@1 on Terminal Bench 1.0, Terminal Bench 2.0, and Terminal Bench Pro, respectively, while smaller variants also consistently improve over their corresponding base models. Additionally, we build \texttt{LiteCoder-Terminal-RL}, a collection of 602 executable terminal environments materialized with \texttt{LiteCoder-Terminal-Gen}, to support verifier-grounded rollouts and trajectory-level preference optimization. Applying DMPO on \texttt{LiteCoder-Terminal-RL} further improves the 4B SFT model on Terminal Bench 2.0 and Terminal Bench Pro, showing that synthesized executable environments can provide useful preference-learning signals beyond supervised fine-tuning.

The contributions of this paper can be summarized as:

\begin{itemize}[leftmargin=1.5em]
\setlength{\leftmargini}{0pt}
    \item We introduce \texttt{LiteCoder-Terminal-Gen}, a zero-dependency synthesis framework that autonomously generates tailored terminal environments, tasks, and robust scoring oracles from scratch to systematically address specific agent capability deficits.
    \item We open-source the \texttt{LiteCoder-Terminal} agent alongside the \texttt{LiteCoder-Terminal-SFT} dataset with 11,255 expert interaction trajectories and the \texttt{LiteCoder-Terminal-RL} dataset with 602 executable and verifiable terminal environments, providing the community with a critical, large-scale resource to overcome the scarcity of system-level training data.
    \item We demonstrate that training on our synthesized data improves terminal-agent performance across Terminal Bench 1.0, Terminal Bench 2.0, and Terminal Bench Pro; supervised fine-tuning yields strong gains across model scales, while DMPO on \texttt{LiteCoder-Terminal-RL} provides further improvements for the 4B SFT model on the harder Terminal Bench 2.0 and Pro benchmarks.
\end{itemize}

\section{Related Work}
\label{gen_inst}
\paragraph{Scaling Environments for Long-horizon Terminal Tasks.} Despite the significant progress frontier models have achieved on repository-level software engineering tasks \citep{jimenez2023swe}, mastering the terminal beyond pure code maintenance remains an open challenge, because these tasks require agents to manage latent system states and interpret raw textual feedback over lengthy context windows. While recent benchmarks like Terminal-Bench \citep{merrill2026terminal} have established rigorous evaluation protocols, the field lacks a scalable method to generate diverse, execution-ready training environments. We also note that throughout the iteration cycle and multiple open-source releases of our dataset\footnote{December, 2025--present}, several high-value data resources have emerged within the field, including the concurrent works by~\citet{pi2026data}, ~\citet{zhu2026termigen} and
~\citet{wu2026largescaleterminalagentictrajectory}. It is precisely these efforts that have driven the collective advancement of the open-source community.

\paragraph{Language Agents Training.} Large-scale agentic training has become a central theme in recent frontier models~\citep{team2025kimi, zeng2026glm, deepseekai2026deepseekv4}. However, the methodologies employed by even the most prominent "open-source" models remain largely opaque; the core training data and recipes lack public implementations. While some existing works~\citep{cai2025nex} have released subsets of agentic data, they generally lack coverage of terminal-task scenarios. Concurrently, recent efforts such as OpenThoughts-Agent~\citep{openthoughts-agent} have attempted to bridge the training gap by converting existing datasets like NL2Bash and InferredBugs into interactive formats. However, these tasks are primarily focused on short-sequence command generation or isolated bug-fixing, which may lack the latent long-horizon supervision signals necessary for complex system manipulation.

\section{LiteCoder-Terminal-Gen: Terminal Tasks Generation at Scale}
\label{sec:LiteCoder-Terminal_gen}

To overcome the scarcity of environment-grounded training data, we introduce \texttt{LiteCoder-Terminal-Gen}, a zero-dependency synthesis pipeline designed to construct executable and verifiable terminal task environments from scratch. Given a high-level domain specification, the framework autonomously generates candidate tasks and materializes them into fully interactive environments.

\subsection{Domain-to-Task Generation}
\begin{figure*}[t]
  \centering
  \includegraphics[width=0.96\textwidth]{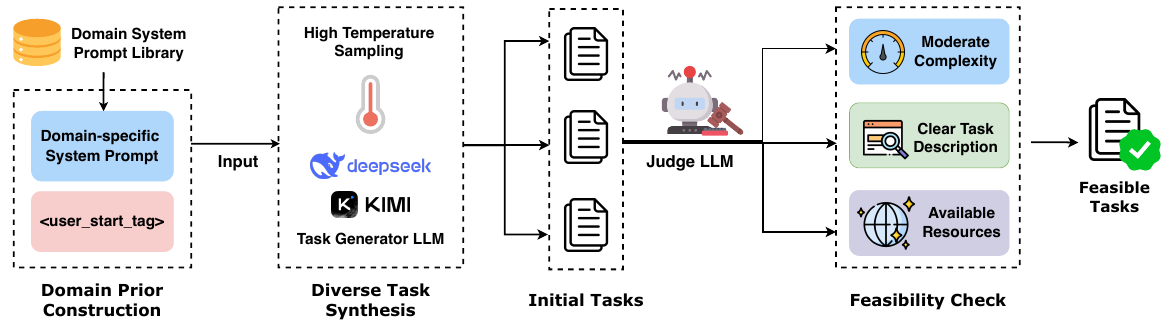}
  \caption{Overview of the domain-to-task generation stage in \texttt{LiteCoder-Terminal-Gen}. Starting from a target terminal domain, we construct a domain-specific prompt prefix that elicits a raw task description from the LLM, then apply a feasibility check that retains only tasks satisfying criteria such as moderate complexity, a clear task description, and available resources.}
  \label{fig:domain_to_task_generation}
\end{figure*}

We begin by specifying a set of terminal domains that cover a broad range of terminal tasks, including \texttt{AI\&ML}, \texttt{build tools}, \texttt{data science}, \texttt{networking}, \texttt{security}, \texttt{system administration}, \texttt{version control}, \texttt{coding}, \texttt{scientific computing}, and \texttt{games}. We then generate tasks conditioned on each domain using a Magpie-like~\citep{xu2025magpie} LLM sampling strategy, as illustrated in Figure~\ref{fig:domain_to_task_generation}. Instead of relying on existing user queries or reference web resources (e.g., GitHub / Stack Overflow), we design domain-specific system prompts to steer task synthesis toward each target domain.

Specifically, we leverage the autoregressive nature of aligned LLMs by completing a partial conversation context. We directly concatenate a pre-query template identifier (e.g., $\texttt{<user\_start>}$) to this system prompt, without supplying any actual user input. This trailing identifier effectively prompts the model into the role of the user, generating the missing turn. By controlling the system prompt, we steer the model to synthesize a specific, high-quality task query that aligns with the target domain. This is immediately followed by a feasibility check that retains only tasks satisfying a set of criteria, including moderate complexity, a clear task description, and available resources.

\subsection{Executable Environment Synthesis}
\label{sec:env_synth}

\begin{figure*}[t]
  \centering
  \includegraphics[width=\textwidth]{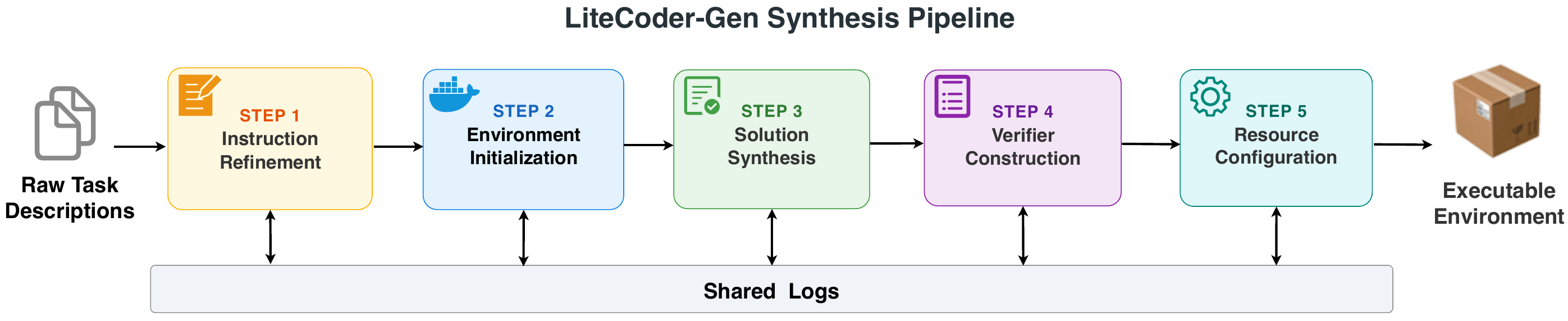}
  \caption{Overview of the executable environment synthesis stage in \texttt{LiteCoder-Terminal-Gen}. Each raw task description is expanded into a full executable environment through a five-stage sequential pipeline---instruction refinement, environment materialization, solution synthesis, verifier crafting, and config derivation---with every stage reading from a shared log directory to prevent inter-stage drift.}
  \label{fig:env_synth}
\end{figure*}

Although the raw task descriptions sampled from the previous step are semantically rich, they are not directly executable. While they effectively capture the user's intent, they often lack the concrete file layouts, background artifacts, expected outputs, and verifiable success criteria essential for an interactive terminal environment. To turn such descriptions into training environments, \texttt{LiteCoder-Terminal-Gen} synthesizes each task through a five-stage sequential pipeline, as illustrated in Figure~\ref{fig:env_synth}. The pipeline progressively refines the task, initializes the environment, synthesizes a reference solution, constructs a verifier, and derives the final configuration. Crucially, each generation stage is explicitly conditioned on the cumulative execution trace of all prior steps. This sequential grounding ensures causal consistency throughout the synthesis process, preventing logical errors—such as a verifier evaluating non-existent artifacts.

We adopt the Harbor task format~\citep{Shaw_Harbor_Framework_2025} as our unified interface for specifying executable tasks and collecting agent trajectories. Each task is organized as a self-contained directory with five key components: (1) an instruction file detailing the natural-language goal; (2) an environment setup, typically a Dockerfile and input artifacts; (3) a reference solution to validate the task design; (4) test scripts that evaluate the agent's success and record rewards; and (5) a configuration file specifying metadata and execution resource limits.

\paragraph{Stage 1: Instruction Refinement.} The \emph{Refiner Agent} takes the raw task description produced by the domain-to-task generation stage and rewrites it into a testable specification. Two constraints are enforced: (i) all input and output files must be bound to concrete absolute paths under the fixed working directory \texttt{/app} (e.g., \texttt{/app/input.json}, \texttt{/app/output.csv}), removing ambiguity that downstream verifiers cannot recover from; (ii) output formats are specified using deterministic schemas (e.g., JSON keys, CSV columns, and floating-point precision). The agent is explicitly prompted to not leak any solution hints, implementation strategies, or related test cases into the final instruction.

\paragraph{Stage 2: Environment Initialization.} Given the refined instruction, the \emph{Environment Agent} produces the \texttt{environment/} directory containing (a) a \texttt{Dockerfile} and (b) all input artifacts referenced by the instruction. Rather than authoring a Dockerfile from scratch, the agent extends a base template supplied by the pipeline. The template pins \texttt{Ubuntu 24.04} as the base OS and pre-installs the necessary runtime dependencies. The prompts are tuned to ensure that the prepared dependencies do not trivially simplify the task.
 
\paragraph{Stage 3: Solution Synthesis.} The \emph{Solver Agent} is tasked with producing a complete, executable \texttt{solution/solve.sh} that satisfies every constraint in \texttt{instruction.md}. The resulting artifact plays two roles. First, it acts as a constructive solvability check: the existence of a runnable \texttt{solve.sh} certifies that the task is actually achievable by an agent; second, it provides a reference point for checking whether the Stage 4 verifier behaves as intended.

\paragraph{Stage 4: Verifier Generation.} The \emph{Verifier Agent} generates two files. The first, \texttt{tests/test.sh}, is mostly template code that serves as the verifier's entry point and writes a binary reward to \texttt{/logs/verifier/reward.txt}. The second, \texttt{tests/test\_outputs.py}, contains the actual test logic as a pytest suite. Because this suite is generated after the oracle solution, it can easily overfit to that specific implementation and reject other valid solutions. To ensure the quality of the verifier (rejecting lazy solutions while accepting legitimate variants), we prompt the agent to execute a mandatory four-phase adversarial iteration before finalizing each assertion:
\begin{itemize}[leftmargin=1.5em]
\setlength{\itemsep}{0pt}
\item \textbf{Draft}: Write an initial validation check based on the task specification.  
\item \textbf{Attack}: Simulate a lazy student that emits an empty file, incorrect data, or a hardcoded dummy payload. If any of these pass, the assertion is too weak.
\item \textbf{Refine}: Simulate an expert agent that uses a different implementation strategy while still satisfying the task specification. If the assertion false-rejects, it is over-specified.
\item \textbf{Finalize}: Write the final robust version based on the preceding attack and refinement steps.
\end{itemize}

\paragraph{Stage 5: Resource Configuration.} The final \emph{Config Agent} reads all four upstream artifacts and emits \texttt{task.toml}, which declares the verifier, agent, and build timeouts, CPU, memory, and storage quotas needed by the task. Resource requirements are estimated by jointly considering the generated artifacts from earlier stages.

Each stage terminates in a lightweight existence check for its expected outputs (\texttt{instruction.md}, \texttt{environment/Dockerfile}, \texttt{solution/solve.sh}, at least one \texttt{tests/test*.\{py,sh\}}, and \texttt{task.toml}). Any stage that fails this check triggers a retry mechanism.

\subsection{Trajectory Collection}

To create the SFT dataset, we collect trajectories with Harbor using MiniMax M2~\citep{minimax2025m2} and M2.1~\citep{minimax2025m21} as teacher models across multiple agent scaffolds, including Terminus\footnote{\url{https://www.tbench.ai/terminus}}, Claude Code~\citep{claude_code_2025}, and OpenHands~\citep{wang2024openhands}. Each run produces a terminal interaction trajectory containing the agent's reasoning, command actions, and environment observations, thereby capturing the thought-action-observation loops required for long-horizon terminal problem solving.

\subsection{Trajectory Filtering}

Quality control is critical for synthetic data. We employ an LLM judge to rigorously filter trajectories based on four behavioral dimensions, retaining only those that demonstrate robust task-solving behavior:

\begin{itemize}[leftmargin=1.5em]
\item \textbf{Adaptability:} We check if the agent can change its plan when it hits an error. We remove trajectories where the agent gets stuck in a loop (repeating the exact same command) or just makes tiny syntax tweaks without changing its overall approach. A good trajectory shows the agent understanding the cause of the error and switching to a new tool or strategy.
\item \textbf{Groundedness:} We make sure the agent pays attention to actual results rather than making things up. We drop trajectories if the agent ignores error messages, assumes it succeeded without actually verifying, or forgets the mistakes it just made.
\item \textbf{Persistence:} We want to see the agent keep trying. We filter out examples where the agent gives up right away when it faces a problem (like a "command not found" error), rather than looking for a reasonable workaround.
\item \textbf{Explicit Refusal:} We simply exclude any trajectories where the agent flat-out refuses to do the task, ensuring our final dataset remains helpful and cooperative.
\end{itemize}

\subsection{Data Decontamination}

We perform strict $N$-gram overlap filtering between our generated task instructions and the test queries in the evaluation benchmarks. Following common practices~\citep{brown2020language, guha2025openthoughts}, we extract all 13-grams from the Terminal Bench datasets and filter out any potentially overlapping tasks. We refer to the remaining decontaminated dataset as \texttt{LiteCoder-Terminal-SFT}.

\section{Data Analysis}

\subsection{Dataset Statistics}

\begin{figure}[t]
  \centering
  \includegraphics[width=\linewidth]{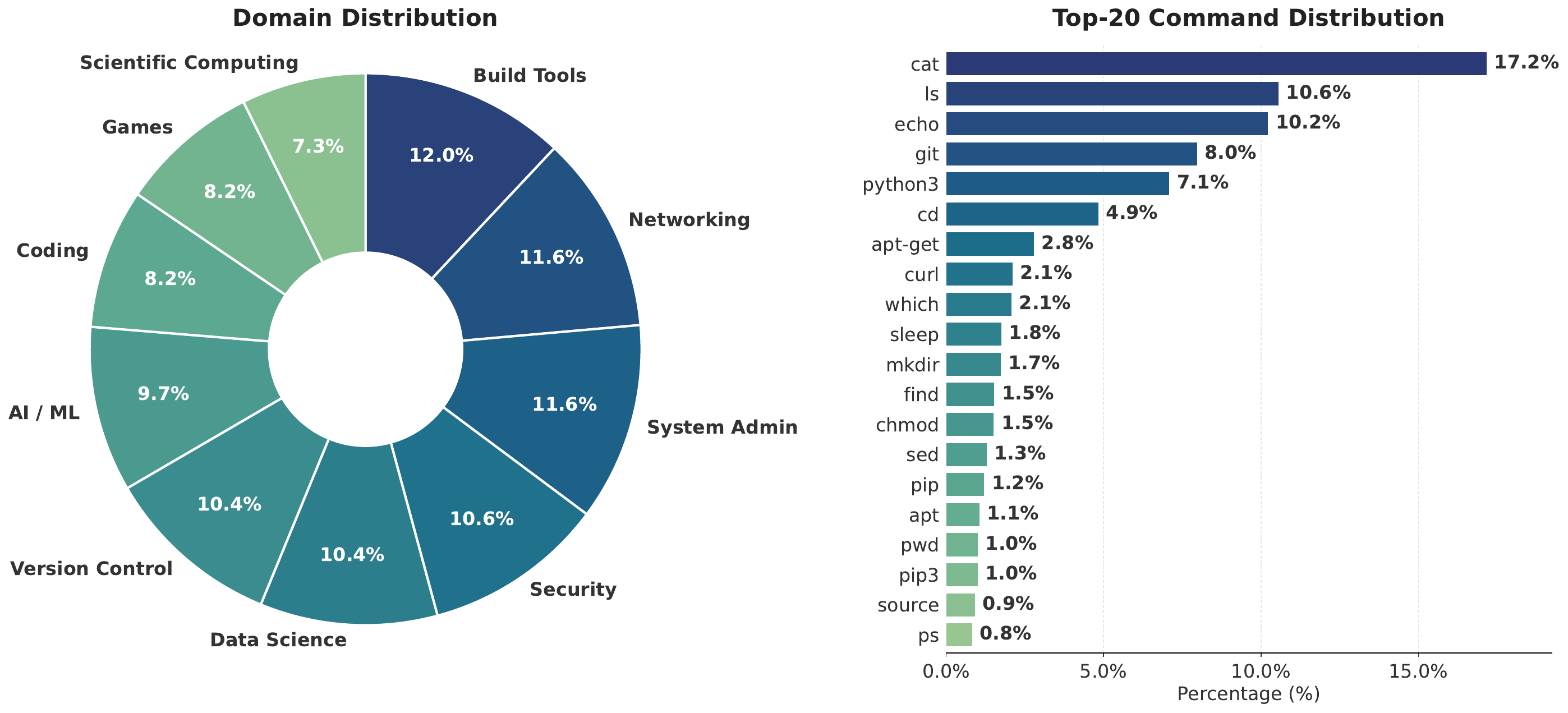}
  \caption{Domain distribution and the top-20 invoked commands in the \texttt{LiteCoder-Terminal-SFT} dataset.}
  \label{fig:data_stats}
\end{figure}

The \texttt{LiteCoder-Terminal-SFT} dataset comprises \textbf{11,255 expert trajectories} spanning \textbf{10 task categories}, with an average of \textbf{27.4 turns} per trajectory. Figure~\ref{fig:data_stats} shows the category distribution. Task categories are roughly balanced, with system administration (11.6\%), networking (11.6\%), and build tools (12.0\%) being the largest groups, while scientific computing (7.3\%) is the smallest. The dataset incorporates trajectories from three agent scaffolds: Terminus-2 (86.6\%), OpenHands (7.1\%), and Claude Code (6.3\%).

\subsection{Command Coverage}

To assess whether \texttt{LiteCoder-Terminal-Gen} produces tasks that elicit broad and realistic terminal behavior, we analyze the commands actually executed in the collected expert trajectories.
We tokenize the first command of every keystroke entry across all 11{,}255 expert trajectories and intersect the resulting vocabulary with the tldr-pages curated Linux command index. After this filter, the trajectories invoke \textbf{over 720 distinct real Linux commands}, spanning from very commonly used utilities---file inspection (\texttt{cat}, \texttt{ls}, \texttt{head}, \texttt{tail}, \texttt{wc}, \texttt{find}, \texttt{grep}), source control (\texttt{git}), package management (\texttt{apt}, \texttt{apt-get}, \texttt{dpkg}, \texttt{pip}, \texttt{cargo}), build and language toolchains (\texttt{make}, \texttt{gcc}, \texttt{go}, \texttt{python3}), system administration (\texttt{chmod}, \texttt{ps}, \texttt{systemctl}, \texttt{ufw}, \texttt{su}), networking and security (\texttt{curl}, \texttt{wget}, \texttt{openssl}, \texttt{gpg}, \texttt{nginx})---all the way to rare specialist tools such as \texttt{mongod}, \texttt{kubeadm}, \texttt{grafana-cli}, \texttt{bison}, \texttt{nasm}, and \texttt{lvcreate}. The 20 most frequently invoked commands are shown in Figure~\ref{fig:data_stats}. This broad command usage demonstrates that our domain-driven sampling captures a wide variety of practical terminal tasks, rather than just standard coding workflows.

\section{Experiments}
\label{sec:experiments}

\subsection{Training Setup}

We evaluate \texttt{LiteCoder-Terminal-Gen} through two complementary training paradigms. First, supervised fine-tuning (SFT) on \texttt{LiteCoder-Terminal-SFT} validates trajectory quality. Second, we use Direct Multi-turn Preference Optimization (DMPO)~\citep{shi2024direct} on \texttt{LiteCoder-Terminal-RL} to evaluate the reliability of our synthesized verifiers. Applying standard DPO to multi-turn interactions is mathematically suboptimal because it treats the sequence as a single-step bandit problem, which ignores the changing environmental states. DMPO addresses this flaw by incorporating a discounted state-action occupancy measure. The objective of DMPO is:

$$ \mathcal{L}_{\text{DMPO}} = - \mathbb{E}_{(\tau_w, \tau_l) \sim \mathcal{D}} \left[ \log \sigma \left( \beta \sum_{t=1}^{|\tau_w|} \alpha_{t}^{(w)} \log \frac{\pi_\theta(a_t^w|s_t^w)}{\pi_{\text{ref}}(a_t^w|s_t^w)} - \beta \sum_{t=1}^{|\tau_l|} \alpha_{t}^{(l)} \log \frac{\pi_\theta(a_t^l|s_t^l)}{\pi_{\text{ref}}(a_t^l|s_t^l)} \right) \right] $$

where $s_t$ and $a_t$ represent the state and action sequences at turn $t$. The weight $\alpha_t$ is the normalized discount factor for a trajectory of length $T$:
$ \alpha_t = \frac{\gamma^{t-1}(1 - \gamma^{T-t+1})}{1 - \gamma^T} $. Applying DMPO serves as a rigorous evaluation of the verifiers themselves. If this training improves model performance, it demonstrates that the auto-generated environments provide valid reward signals capable of guiding long-horizon optimization.

To perform DMPO, we construct trajectory-level preference pairs using our synthesized environments. Starting from the \texttt{LiteCoder-Terminal-4b-sft} checkpoint, we sample two independent rollouts for each of the 602 environments in LiteCoder-Terminal-RL. We compute a pass ratio (the fraction of verifier checks satisfied) for each rollout and retain only the environments where the two trajectories yield divergent scores. The higher-scoring trajectory serves as the preferred multi-turn response, while a lower-scoring trajectory from the same environment is selected as the rejected response.

\subsection{Evaluation Setup}

\paragraph{Benchmarks and Scaffolds.}

We evaluate our models on Terminal Bench 1.0\footnote{\url{https://www.tbench.ai/leaderboard/terminal-bench/1.0}}, Terminal Bench 2.0~\citep{merrill2026terminal}, and Terminal Bench Pro~\citep{wang2025let}, reporting the Pass Rate (\%) as the primary metric for model capability. To mitigate variance and obtain robust estimates, the scores for Terminal Bench 1.0 and 2.0 are averaged across four independent runs. Additionally, we report the pass@4 metric. While we deploy Terminus-2 as our default agentic scaffold, we evaluate the Nex-N1 baseline using OpenHands to align with its original technical report and ensure a fair comparison.

\paragraph{Models and Baselines.} Our empirical study leverages representative instruction-tuned backbones: \texttt{Qwen3-\{4B/30B-A3B\}-Instruct}~\citep{yang2025qwen3}, and \texttt{Qwen2.5-Coder-32B-Instruct}~\citep{hui2024qwen2}. We fine-tune each base model on our proposed corpus, yielding the \texttt{LiteCoder-Terminal} models. Comprehensive training details and hyperparameters are deferred to Appendix~\ref{app:training_details}. Furthermore, we include representative SFT baselines, such as \texttt{Qwen3-30B-A3B-Nex-N1}~\citep{cai2025nex} and \texttt{OpenThinker-Agent-v1}~\citep{openthoughts-agent}.

\subsection{Overall Results}

\subsubsection{Effectiveness of LiteCoder-Terminal-SFT}

\begin{table}[t]
  \caption{Terminal task benchmark results at pass@1 and pass@4 (\%). TB-1 / 2 / Pro stands for Terminal Bench-1.0 / 2.0 / Pro.}
  \label{tab:agentic_benchmark_combined}
  \centering
  \footnotesize
  \renewcommand{\arraystretch}{1.2}
  \setlength{\tabcolsep}{4pt}
  \begin{tabular}{@{} l l c c c c c c c @{}}
    \toprule
    \multirow{2}{*}{\textbf{Model}} & \multirow{2}{*}{\textbf{Scaffold}} & \multicolumn{4}{c}{\textbf{Pass@1}} & \multicolumn{3}{c}{\textbf{Pass@4}} \\
    \cmidrule(lr){3-6} \cmidrule(lr){7-9}
    & & \textbf{TB-1} & \textbf{TB-2} & \textbf{TB-Pro} & \textbf{Avg.} & \textbf{TB-1} & \textbf{TB-2} & \textbf{Avg.} \\
    \midrule
    \multicolumn{9}{@{}l}{\textit{Base instruction-tuned models}} \\
    Qwen3-4B-Instruct~\citep{yang2025qwen3}                   & Terminus-2 & $6.25_{\pm 1.77}$  & $1.12_{\pm 1.12}$  & 3.50  & 3.62  & 15.00 & 3.37  & 9.19  \\
    Qwen3-30B-A3B-Instruct~\citep{yang2025qwen3}              & Terminus-2 & $16.56_{\pm 3.29}$ & $5.34_{\pm 1.69}$  & 20.50 & 14.13 & 28.75 & 11.24 & 20.00 \\
    Qwen2.5-Coder-32B-Instruct~\citep{hui2024qwen2}           & Terminus-2 & $12.19_{\pm 3.08}$ & $4.49_{\pm 1.72}$  & 13.50 & 10.06 & 20.00 & 8.99  & 14.50 \\
    \addlinespace
    \multicolumn{9}{@{}l}{\textit{Baselines}} \\
    OpenThinker-Agent-v1~\citep{openthoughts-agent} & Terminus-2 & $11.25_{\pm 1.77}$ & $4.49_{\pm 3.18}$  & 19.50 & 11.75 & 25.00 & 10.10 & 17.55 \\
    Qwen3-30B-A3B-Nex-N1~\citep{cai2025nex}         & OpenHands  & $18.44_{\pm 3.13}$ & $12.36_{\pm 2.05}$ & 21.00 & 17.27 & 32.50 & 23.60 & 28.05 \\
    Qwen3-32B-Nex-N1~\citep{cai2025nex}              & OpenHands  & $24.69_{\pm 1.56}$ & $18.54_{\pm 1.95}$ & 30.50 & 24.58 & 35.00 & 26.97 & 30.99 \\
    TerminalTraj-32B~\citep{wu2026largescaleterminalagentictrajectory}        & Terminus-2 & $33.44_{\pm 3.44}$ & $23.88_{\pm 2.95}$ & 30.50 & 29.27 & 45.00 & 37.08 & 41.04 \\
    Nemotron-Terminal-32B\tablefootnote{We note a gap between our reproduced Nemotron-Terminal-32B results and those reported in the original paper. After reviewing publicly reported reproduction results, we find that using a 16x timeout multiplier (i.e., 16 times the default timeout) yields performance close to the original report, whereas a 2x multiplier yields results closer to those we report. For fair comparison, we therefore report metrics without applying any timeout multiplier.}~\cite{pi2026data}   & Terminus-2 & $27.81_{\pm 3.29}$ & $21.35_{\pm 2.75}$ & 37.00 & 28.72 & 46.25 & 35.96 & 41.11 \\
    \addlinespace
    \multicolumn{9}{@{}l}{\textit{Fine-tuned on LiteCoder-Terminal-SFT}} \\
    \rowcolor{cyan!10} LiteCoder-Terminal-4b-sft       & Terminus-2 & $14.69_{\pm 1.20}$ & $4.78_{\pm 1.83}$  & 21.50 & 13.66 & 28.75 & 10.11 & 19.43 \\
    \rowcolor{cyan!10} LiteCoder-Terminal-30b-a3b-sft  & Terminus-2 & $24.38_{\pm 1.61}$ & $12.36_{\pm 2.75}$ & 31.50 & 22.75 & 40.00 & 23.60 & 31.80 \\
    \rowcolor{cyan!10} LiteCoder-Terminal-32b-sft      & Terminus-2 & ${29.06}_{\pm 4.18}$ & ${18.54}_{\pm 3.40}$ & 34.00 & 27.20 & 45.00 & 30.34 & 37.67 \\
    \bottomrule
  \end{tabular}
\end{table}

Table~\ref{tab:agentic_benchmark_combined} shows that training on \texttt{LiteCoder-Terminal-SFT} consistently improves terminal-agent performance across model scales and benchmarks: the fine-tuned \texttt{LiteCoder-Terminal} models outperform their corresponding backbones across all three scales. Specifically, on Terminal Bench 1.0, the 4B, 30B-A3B, and 32B variants surpass their respective base models by 8.44, 7.82, and 16.87 absolute percentage points. On Terminal Bench 2.0, the improvements are even more pronounced: the 4B and 32B models achieve more than four-fold increases in pass rate, while the 30B-A3B model more than doubles the performance of its base counterpart. Most notably, on Terminal Bench Pro---which enforces balanced domain distribution---the 32B variant achieves 34.00\% pass@1, while the 30B-A3B and 4B variants reach 31.50\% and 21.50\%, respectively. These gains indicate that our synthesized expert trajectories provide effective training signals for the command-line skills needed in rigorous terminal benchmarks.

\subsubsection{Comparison with Baseline Models}

\texttt{LiteCoder-Terminal} achieves competitive performance using a substantially smaller training set than existing baselines, consistently outperforming or matching the corresponding \texttt{Nex-N1}~\citep{cai2025nex} models at both 30B and 32B scales across Terminal Bench 1.0, 2.0, and Pro. Crucially, these results highlight the efficacy and data efficiency of our synthesis paradigm: while large-scale datasets like \texttt{TerminalTraj}~\citep{wu2026largescaleterminalagentictrajectory} (50.7K trajectories) and \texttt{Nemotron-Terminal}~\citep{pi2026data} (490.5K) rely heavily on mining existing human-curated repositories, \texttt{LiteCoder-Terminal-Gen} autonomously synthesizes both the executable environments and complex interaction tasks entirely from scratch. Despite utilizing up to 43.6$\times$ fewer trajectories (11.2K), our 32B model remains highly competitive, surpassing \texttt{Nemotron-Terminal} on Terminal Bench 1.0, achieving top-tier results on Terminal Bench Pro, and maintaining a narrow gap in average pass@1 performance. Ultimately, this demonstrates that zero-dependency environment synthesis provides a highly data-efficient supervision signal for enhancing terminal-agent capabilities.

\subsubsection{Effectiveness of LiteCoder-Terminal-RL}

\begin{wraptable}{r}{0.48\linewidth}
  \centering
  \vspace{-1.8em}
  \caption{Effect of DMPO on pass@1 (\%) for Qwen3-4B-Instruct.}
  \label{tab:dmpo_results}
  \vspace{0.4em}
  \footnotesize
  \renewcommand{\arraystretch}{1.2}
  \setlength{\tabcolsep}{6pt} 
  \begin{tabular}{@{} l c c c c @{}}
    \toprule
    \textbf{Method} & \textbf{TB-1} & \textbf{TB-2} & \textbf{TB-Pro} & \textbf{Avg.} \\
    \midrule
    SFT Baseline & \textbf{14.69} &  4.78 & 21.50 & 13.66 \\
    + DMPO       & 14.38 & \textbf{6.10} & \textbf{23.00} & \textbf{14.49} \\
    \bottomrule
  \end{tabular}
  \vspace{-0.8em}
\end{wraptable}

As shown in Table~\ref{tab:dmpo_results}, applying DMPO improves average performance over the SFT baseline, with gains on the harder Terminal Bench 2.0 and Terminal Bench Pro benchmarks. It increases pass@1 from 4.78\% to 6.10\% on Terminal Bench 2.0, and from 21.50\% to 23.00\% on Terminal Bench Pro. This indicates that LiteCoder-Terminal-RL environments are highly coherent; the verifiers provide trustworthy, correctness-grounded optimization targets capable of steering long-horizon agent behavior beyond what is achievable through behavioral cloning alone.

\subsection{Detailed Analysis}

\subsubsection{Domain Ablation}
To understand the contribution of individual domains to overall capability, we perform a leave-one-domain-out ablation on our balanced subset. We fine-tune the \texttt{Qwen3-4B-Instruct} base model on reduced mixtures and compare performance against the full dataset (Table \ref{tab:data_ablation}). Removing any single domain yields only a modest average degradation, indicating a distributed reliance on diverse task types rather than a single critical domain. Notably, removing domains like "games" or "security" triggers the sharpest overall declines ($\downarrow$ 2.50 and $\downarrow$ 2.15, respectively), suggesting that these domains present highly challenging scenarios that rigorously test the model's capacity for edge-case reasoning and complex dependency management—capabilities that transfer broadly across the benchmark suite.

\begin{table}[t]
  \caption{Domain ablation results. We report the performance after removing each domain from the training mixture. Domains are sorted by their impact on the average score ($\Delta$ Avg.), highlighting their relative importance to the model's overall capability.}
  \label{tab:data_ablation}
  \centering
  \footnotesize
  \renewcommand{\arraystretch}{1.2}
  \setlength{\tabcolsep}{5pt}
  \begin{tabular}{@{} l cccc | c @{}}
    \toprule
    \multirow{2}{*}{\textbf{Removed Domain}} & \multicolumn{4}{c|}{\textbf{Absolute Performance (\%)}} & \textbf{Impact} \\
    \cmidrule{2-6}
    & \textbf{TB-1} & \textbf{TB-2} & \textbf{TB-Pro} & \textbf{Avg.} & $\Delta$ \textbf{Avg.} \\
    \midrule
    \textbf{Full Data (Baseline)} & 14.69 & 4.78 & \textbf{21.50} & \textbf{13.66} & - \\
    \midrule
    w/o games              & 11.88 & 5.60 & 16.00 & 11.16 & \color{red}{$\downarrow$ 2.50} \\
    w/o security           & 11.25 & 4.78 & 18.50 & 11.51 & \color{red}{$\downarrow$ 2.15} \\
    w/o version control    & 12.81 & 3.93 & 19.50 & 12.08 & \color{red}{$\downarrow$ 1.58} \\
    w/o coding             & 13.44 & 5.30 & 17.50 & 12.08 & \color{red}{$\downarrow$ 1.58} \\
    w/o data science       & 13.13 & 6.18 & 17.50 & 12.27 & \color{red}{$\downarrow$ 1.39} \\
    w/o ai\_ml             & 10.94 & 5.62 & 20.50 & 12.35 & \color{red}{$\downarrow$ 1.31} \\
    w/o system admin       & 14.06 & \textbf{7.86} & 16.00 & 12.64 & \color{red}{$\downarrow$ 1.02} \\
    w/o networking         & 13.75 & 5.90 & 19.50 & 13.05 & \color{red}{$\downarrow$ 0.61} \\
    w/o build tools        & \textbf{16.88}& 4.50 & 18.50 & 13.29 & \color{red}{$\downarrow$ 0.37} \\
    w/o scientific comp.   & 14.38 & 4.78 & 21.00 & 13.39 & \color{red}{$\downarrow$ 0.27} \\
    \bottomrule
  \end{tabular}
\end{table}

\subsubsection{Test-Time Scaling}
\label{sec:ttscale}

\begin{figure}[t]
  \centering
  \includegraphics[width=\linewidth]{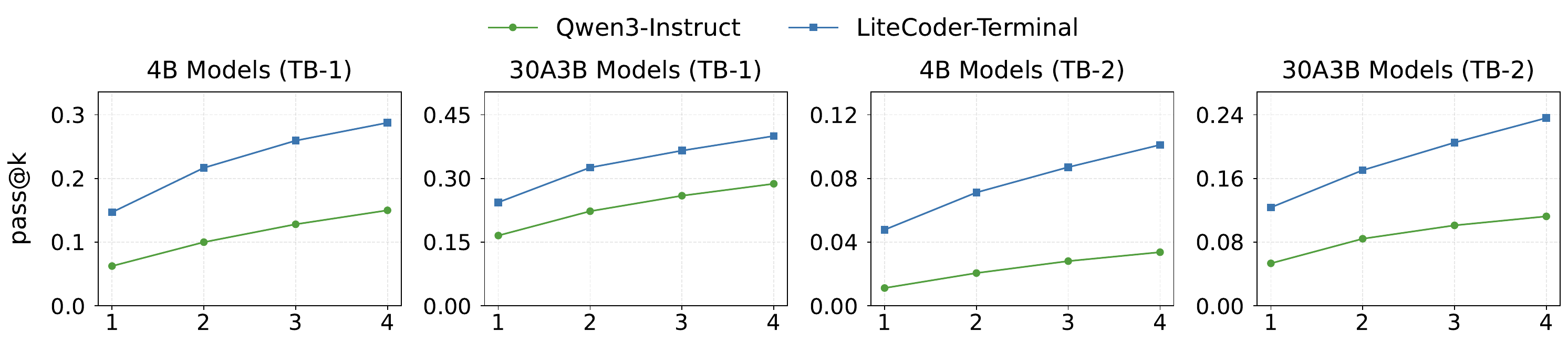}
  \caption{Pass@$k$ across different sampling budgets $k$ on Terminal Bench 1.0 and 2.0 for the 4B and 30B-A3B scales. Green: base \texttt{Qwen3-Instruct}; blue: \texttt{LiteCoder-Terminal} fine-tuned on SFT trajectories.}
  \label{fig:ttscale}
\end{figure}

We analyze test-time scaling behavior by tracking pass@$k$ performance. Models lacking robust terminal-solving capabilities tend to plateau quickly, as repeated attempts merely reproduce identical failure modes. Figure~\ref{fig:ttscale} illustrates that \texttt{LiteCoder-Terminal} models possess a distinctly stronger capacity to exploit increased sampling budgets. On Terminal Bench 1.0, the 30B-A3B variant scales from 24.4\% at $k=1$ to 40.0\% at $k=4$, a 15.6-point gain that outpaces the base model's trajectory. This steep scaling curve on both TB-1 and TB-2 indicates that SFT on our dataset not only improves the single-attempt pass rate (pass@1) but fundamentally enhances the agent's latent capacity to explore and eventually uncover correct execution paths.

\subsubsection{Cross-Task Generalization to SWE-bench}

\begin{wrapfigure}{r}{0.42\linewidth}
  \centering
  \vspace{-1.8em}
  \includegraphics[width=\linewidth]{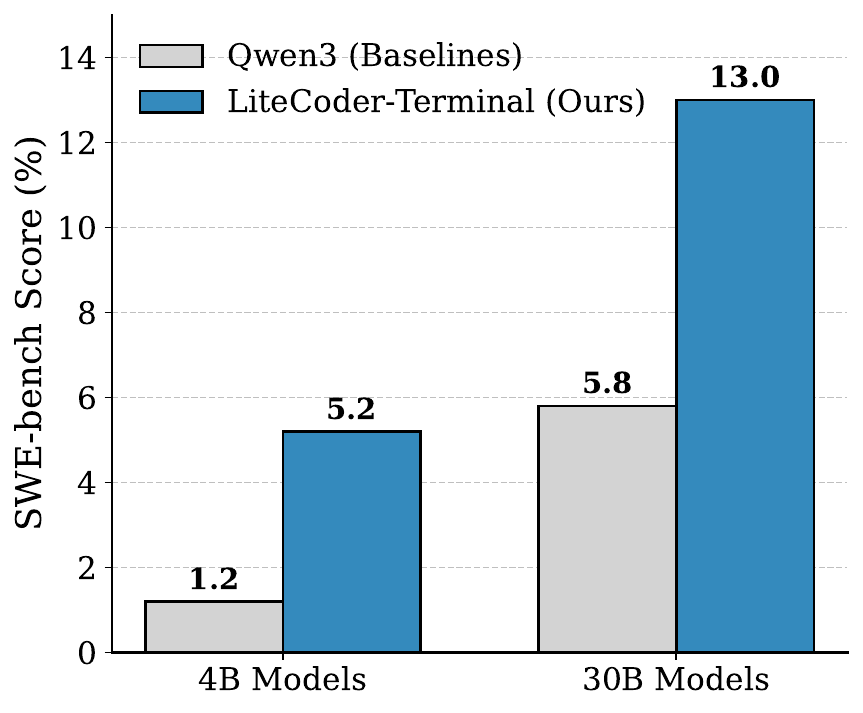}
  \caption{Cross-task evaluation on SWE-bench.}
  \label{fig:cross_task_swe}
  \vspace{-1.5em}
\end{wrapfigure}

To examine whether the learned terminal-agent behaviors carry over to software engineering tasks, we additionally evaluate our trained models on SWE-bench.

Empirical results demonstrate that the terminal interaction capabilities acquired via \texttt{LiteCoder-Terminal-SFT} successfully generalize to SWE-bench. As illustrated in Figure~\ref{fig:cross_task_swe}, the fine-tuned models consistently outperform their base counterparts: the resolution rate of the 4B model improves from 1.2\% to 5.2\%, while the 30B-A3B model exhibits a substantial increase from 5.8\% to 13.0\%. Although our data pipeline is not explicitly optimized for SWE-bench, these findings provide evidence that modeling long-horizon terminal trajectories broadly enhances repository-level software engineering workflows within the same scaffold.

\section{Discussion}

\paragraph{Conclusion.}
In this paper, we introduced LiteCoder-Terminal-Gen, a zero-dependency pipeline for synthesizing executable terminal-agent environments. By replacing source-dependent task mining with target-driven synthesis, our framework ensures accurate and scalable coverage of long-horizon command-line skills. Models fine-tuned on our synthesized SFT trajectories consistently outperform their backbones across the Terminal Bench suite, while applying DMPO on our verifier-grounded RL environments unlocks further performance gains on highly complex tasks. These results systematically demonstrate that fully synthetic, executable environments offer a scalable supervision signal for mastering real-world terminal interactions.

\paragraph{Limitations.}\label{sec:limitations}
We acknowledge two key limitations that outline future directions. First, because task instructions are produced via LLM completion, the resulting task distribution inherits biases from the generator model. Second, all environments are instantiated with Ubuntu-based Docker images and predominantly exercise GNU/Linux utilities; extending the pipeline to other Linux distributions and operating systems could help agents move beyond fixed environment assumptions and improve generalization, and remains a direction for future work.

\bibliographystyle{unsrtnat}
\bibliography{neurips_2026}


\appendix

\section{Domain-to-Task Generation Prompt}
\label{app:magpie_prompt}

Below is an example of the domain-specific system prompt used in the Magpie-style active sampling stage (Section~\ref{sec:LiteCoder-Terminal_gen}). This prompt steers the LLM to synthesize task queries for the Data Science domain.

\begin{tcolorbox}[
  colback=cyan!5!white,
  colframe=cyan!60!blue,
  coltitle=white,
  fonttitle=\bfseries,
  title=Domain-specific system prompt (Data Science),
  top=6pt, bottom=6pt, left=8pt, right=8pt
]
\small
Now you are provided with a Ubuntu Docker container. Your task is to solve complex, multi-turn problems by interacting with the terminal.

\medskip
Note that the task descriptions are concise, but the execution is technically demanding.

\medskip
The problem description will follow the following format:

\medskip
\#\# Task Title: [A concise, descriptive title]

\medskip
\textbf{Domain Focus:} Data Science \& Data Processing

\medskip
\textbf{Objective:} [Summarize the final goal of this data science and data processing task in one sentence.]

\medskip
\textbf{Scenario:} [Describe a brief background or context to make the task realistic and engaging.]

\medskip
\textbf{Todo Checklist:}\\
- [ ] Todo 1:\\
- [ ] Todo 2:\\
- [ ] Todo 3:\\
- [ ] Todo 4:\\
- [ ] Todo 5:\\
- ... (continue as needed; 6--10 items total)
\end{tcolorbox}

Each domain uses a prompt with the same structure but with its domain focus replaced accordingly (e.g., ``Networking \& Security'', ``System Administration'', ``AI \& ML'').

\section{Task Filtering Prompt}
\label{app:filtering_prompt}

Below is the prompt used to filter generated terminal tasks before executable environment synthesis. The filter rejects tasks that are infeasible for an autonomous agent to complete in a CPU-only, single-machine Docker environment within a reasonable timeframe.

\begin{tcolorbox}[
  colback=cyan!5!white,
  colframe=cyan!60!blue,
  coltitle=white,
  fonttitle=\bfseries,
  title=Task Filtering,
  breakable
]
\small
You are an Expert DevOps and AI Capability Evaluator.

\medskip
\noindent
Your job is to analyze a generated \textbf{Terminal/Command-Line Task} and determine if it is FEASIBLE for an autonomous AI agent to execute within a \textbf{CPU-only, single-machine Docker environment} in a reasonable timeframe (e.g., $<$ 1 hour) by interacting with the terminal.

\medskip
\noindent
\textbf{Criteria for infeasible tasks (reject these):}
\begin{enumerate}[leftmargin=1.5em]
\item \textbf{Extreme Complexity / Hallucination:} Tasks that are wildly unrealistic or require massive engineering teams.
\begin{itemize}[leftmargin=1.5em]
\item Example: ``Write a GPT-4 level LLM from scratch in C++.''
\item Example: ``Create a full operating system kernel overnight.''
\end{itemize}

\item \textbf{Vague / Ambiguous:} Instructions with no clear success criteria.

\item \textbf{Unavailable Resources:} Tasks that depend on missing hardware (GPUs, physical peripherals) or require external authentication.

\item \textbf{Any other tasks that you deem unreasonable or impractical based on your expert judgment.}
\end{enumerate}

\medskip
\noindent
You must respond in strict JSON format:

\medskip
\noindent\texttt{\{}\\
\texttt{\ \ ``feasible'': boolean,}\\
\texttt{\ \ ``reason'': ``Explanation of why it is accepted or rejected.'',}\\
\texttt{\ \ ``difficulty'': ``Easy/Medium/Hard/Impossible''}\\
\texttt{\}}
\end{tcolorbox}

\section{Environment Synthesis Pipeline Prompts}
\label{app:pipeline_prompts}

Below are the prompts used in the five-stage executable environment synthesis pipeline described in Section~\ref{sec:LiteCoder-Terminal_gen}. Each stage agent reads the shared \texttt{agent\_logs/} directory containing outputs from all preceding stages.

\begin{tcolorbox}[
  colback=cyan!5!white,
  colframe=cyan!60!blue,
  coltitle=white,
  fonttitle=\bfseries,
  title=Instruction Refinement,
  breakable
]
\small
You are an expert at creating testable programming tasks for benchmark evaluation.

\medskip
\noindent
I have a task description that needs to be enhanced with specific testable constraints.

\medskip
\noindent
\textbf{Your job:}

\begin{enumerate}[leftmargin=1.5em]
\item \textbf{Analyze the task} to understand: high-level goal and requirements; programming language and tools needed; expected inputs and outputs; how to make it more testable.

\item \textbf{Transform it into testable format} with specific constraints: clear implementation requirements (functions, classes, specific input/output specifications with concrete file paths, e.g., \texttt{/app/input.json}); data structure handling requirements; technical stack specifications; output format requirements (JSON structure, CSV format, etc.).

\item \textbf{Structure instruction.md} clearly: brief task description (1--2 sentences); technical requirements (language, input and output files); input/output specifications with examples; data format specifications (with precise details); edge cases and error handling (if applicable).
\end{enumerate}

\medskip
\noindent
\textbf{Important:} Use \texttt{/app} as the working directory. Make requirements specific and testable. Avoid leaking test cases or any solution guidance in the task description.
\end{tcolorbox}

\begin{tcolorbox}[
  colback=cyan!5!white,
  colframe=cyan!60!blue,
  coltitle=white,
  fonttitle=\bfseries,
  title=Environment Materialization,
  breakable
]
\small
You are an expert at creating Docker environments for terminal tasks.

\medskip
\noindent
I have a structured task with detailed \texttt{instruction.md}. Your job is to:

\begin{enumerate}[leftmargin=1.5em]
\item \textbf{Analyze the task requirements} to determine what test data files are required as inputs.

\item \textbf{Create \texttt{environment/} directory structure}: \texttt{environment/[data files]} --- any input test data files mentioned in \texttt{instruction.md}; \texttt{environment/Dockerfile} --- container environment based on the base image template.

\item \textbf{Dockerfile requirements}: Start with a fixed base image configuration (Ubuntu 24.04 with tmux, asciinema, uv, Python 3.13, OpenHands, and Claude Code pre-installed). After the base setup, add task-specific configuration: set \texttt{WORKDIR}, \texttt{COPY} test data files to their required locations.

\item \textbf{Test data files}: Create realistic sample data files mentioned in \texttt{instruction.md}. Files should be small but representative. Match exact specifications from \texttt{instruction.md}.
\end{enumerate}

\medskip
\noindent
\textbf{Important:} Do NOT install any additional packages; rely solely on the base image configuration.
\end{tcolorbox}

\begin{tcolorbox}[
  colback=cyan!5!white,
  colframe=cyan!60!blue,
  coltitle=white,
  fonttitle=\bfseries,
  title=Solution Generation,
  breakable
]
\small
You are an expert programmer who creates reference solutions for benchmark tasks.

\medskip
\noindent
I have a task with \texttt{instruction.md} and \texttt{environment/}. Your job is to create a complete reference solution.

\medskip
\noindent
\textbf{Solution structure:} Create \texttt{solution/solve.sh} that must start with a benchmark canary string (a fixed GUID marker for downstream decontamination audits), implement all requirements from \texttt{instruction.md}, handle all edge cases, follow exact output format specifications, and use \texttt{/app} paths.

\medskip
\noindent
\textbf{Quality standards:} Complete working implementation; handles all edge cases; uses appropriate libraries; clean, efficient code; proper input/output handling.

\medskip
\noindent
Previous agent logs in \texttt{agent\_logs/} provide context on how \texttt{instruction.md} was enhanced (Step~1) and what environment and test data were created (Step~2).
\end{tcolorbox}

\begin{tcolorbox}[
  colback=cyan!5!white,
  colframe=cyan!60!blue,
  coltitle=white,
  fonttitle=\bfseries,
  title=Test Crafting,
  breakable
]
\small
You are an expert at creating test suites for Harbor benchmark tasks. High-quality benchmark tests require multiple rounds of mental iteration to be robust.

\medskip
\noindent
I have a complete task with \texttt{instruction.md}, \texttt{environment/}, and \texttt{solution/}. Your job is to create focused, standard tests that verify core functionality.

\medskip
\noindent
\textbf{Test files:}
\begin{itemize}[leftmargin=1.5em]
\item \texttt{tests/test.sh} --- Entry point that installs dependencies (uv, pytest), runs the test suite, and writes a binary reward to \texttt{/logs/verifier/reward.txt}.
\item \texttt{tests/test\_outputs.py} --- Pytest suite that validates output files, structure, content, and edge cases.
\end{itemize}

\medskip
\noindent
\textbf{Mandatory four-phase adversarial iteration for each assertion:}
\begin{enumerate}[leftmargin=1.5em]
\item \textbf{Draft:} Write the initial validation check.
\item \textbf{Attack:} Simulate a lazy agent that emits an empty file, incorrect data, or a hardcoded dummy payload. If any of these pass, the assertion is too weak.
\item \textbf{Refine:} Simulate an expert agent that uses a different implementation approach but produces correct results. If the assertion false-rejects, it is over-specified.
\item \textbf{Finalize:} Write the robust version based on the preceding attack and refinement steps.
\end{enumerate}
\end{tcolorbox}

\begin{tcolorbox}[
  colback=cyan!5!white,
  colframe=cyan!60!blue,
  coltitle=white,
  fonttitle=\bfseries,
  title=Config Derivation,
  breakable
]
\small
You are an expert at creating Harbor benchmark task configurations.

\medskip
\noindent
I have a complete task with \texttt{instruction.md}, \texttt{environment/}, \texttt{solution/}, and \texttt{tests/}. Your job is to analyze everything and create an accurate \texttt{task.toml} configuration.

\medskip
\noindent
\textbf{Requirements:}
\begin{enumerate}[leftmargin=1.5em]
\item \textbf{Analyze the complete task} to determine: task difficulty (easy/medium/hard based on solution complexity); task category; appropriate technology tags (3--5 tags); time estimates for experts and juniors; resource requirements (CPU, memory, storage).

\item \textbf{Examine all generated files}: \texttt{instruction.md}, \texttt{environment/Dockerfile}, \texttt{solution/solve.sh}, and \texttt{tests/}.

\item \textbf{Create \texttt{task.toml}} declaring verifier, agent, and build timeouts, CPU, memory, and storage quotas.
\end{enumerate}

\medskip
\noindent
\textbf{Guidelines:} Resource allocation ranges from basic tasks (1 CPU, 2048 MB) to ML/build tasks (2--4 CPUs, 4096--8192 MB). Verifier timeout ranges from 360s (simple) to 900s (complex builds). Agent timeout ranges from 1800s (simple) to 3600s (complex).

\medskip
\noindent
Previous agent logs from all four preceding steps are available in \texttt{agent\_logs/} for accurate resource estimation.
\end{tcolorbox}

\section{Trajectory Filtering Prompt}
\label{app:trajectory_filtering_prompt}

Below is the prompt used by the LLM judge to filter collected terminal-agent trajectories according to the behavioral criteria described in Section~\ref{sec:LiteCoder-Terminal_gen}.

\begin{tcolorbox}[
  colback=cyan!5!white,
  colframe=cyan!60!blue,
  coltitle=white,
  fonttitle=\bfseries,
  title=Trajectory Filtering,
  breakable
]
\small
You are an expert analyst of AI Agent behaviors, specializing in evaluating \emph{Agentic Capabilities}.

\medskip
\noindent
Your task is to analyze the conversation trace and diagnose how the agent handles obstacles, errors, or feedback. You must classify the agent's behavior into specific categories regarding its \textbf{Adaptability}, \textbf{Groundedness}, and \textbf{Persistence}.

\medskip
\noindent
\textbf{Evaluation Framework.}

\medskip
\noindent
\textbf{1. Adaptability.} Analyze if the agent gets stuck in loops or fails to pivot strategies.
\begin{itemize}[leftmargin=1.5em, itemsep=2pt, topsep=3pt]
\item \textbf{Mechanical Loop:} Repeating the exact same command after encountering an error. This is the lowest level of failure.
\item \textbf{Rigid Strategy:} Although parameters (like syntax) are slightly modified after an error, the logical path to solve the problem remains unchanged (e.g., constantly looking for a file in the wrong folder, just changing the filename), leading to continued failure. This is also a type of ``soft loop''.
\item \textbf{Strategic Pivot:} Analyzing the cause after an error and switching to a completely different tool or path (e.g., search yields nothing $\rightarrow$ switch to browse; tool error $\rightarrow$ check documentation). This is \textbf{not} a loop.
\end{itemize}

\medskip
\noindent
\textbf{2. Groundedness.} Analyze if the agent fails to be reality-aligned.
\begin{itemize}[leftmargin=1.5em, itemsep=2pt, topsep=3pt]
\item \textbf{Ignoring Feedback:} The tool returns an error, but the agent claims ``Task completed'' in the next step.
\item \textbf{Hallucinated Success:} Assuming a file exists or a state is achieved without tool verification.
\item \textbf{Context Drift:} Forgetting that a certain method was already attempted and failed in previous steps.
\end{itemize}

\medskip
\noindent
\textbf{3. Persistence.} Analyze if the agent gives up on the task prematurely when facing obstacles.
\begin{itemize}[leftmargin=1.5em, itemsep=2pt, topsep=3pt]
\item \textbf{Premature Surrender:} Concluding the task is impossible or ``cannot be completed'' immediately after encountering an environmental limitation (e.g., missing compiler, command not found) without attempting reasonable alternatives or workarounds (e.g., checking for other binaries, verifying paths, or trying fallback methods).
\item \textbf{Note:} The conversation may end abruptly due to external constraints (e.g., timeouts or step limits). Do not penalize the agent for an uncompleted task simply because the log ends. A lack of persistence is found only when the agent voluntarily abandons the task (e.g., actively marking \texttt{task\_complete = true} or declaring the task impossible) while the objective remains unfulfilled. If the agent is actively planning or executing a step when the conversation stops, this is not a failure.
\item \textbf{Note:} If the agent discusses the difficulty in its thought process but eventually decides to try another approach (e.g., ``Wait, let me reconsider...''), this demonstrates good persistence and is not a failure.
\end{itemize}

\medskip
\noindent
\textbf{4. Refusal \& Stoppage.} Analyze if the agent explicitly refuses to proceed with the task.
\begin{itemize}[leftmargin=1.5em, itemsep=2pt, topsep=3pt]
\item \textbf{Explicit Refusal:} The agent states it cannot or will not fulfill the request (e.g., ``I cannot assist with this,'' ``I am unable to generate this content'').
\end{itemize}

\medskip
\noindent
\textbf{Important:} Strictly ignore all JSON formatting-related deviations. Examples include incorrect field ordering (since JSON is unordered) or the presence of conversational text surrounding the JSON object. Even if the agent persists in these formatting habits despite warnings, this is a stylistic choice and must \textbf{not} be classified as \textit{Rigid Strategy} or a capability failure. Focus solely on the logical problem-solving actions contained within the data.

\medskip
\noindent
The judge must respond in strict JSON format with the fields \texttt{reason}, \texttt{has\_adaptability\_failure}, \texttt{has\_groundedness\_failure}, \texttt{has\_persistence\_failure}, and \texttt{triggered\_refusal}.
\end{tcolorbox}

\section{Training Details}
\label{app:training_details}

All \texttt{LiteCoder-Terminal} models are fine-tuned using the AutoAlign framework with DeepSpeed ZeRO-3 parallelism on 8 GPUs per node. We use the AdamW optimizer with a learning rate of $5 \times 10^{-6}$, a cosine learning rate scheduler with a warmup ratio of 0.04, and a weight decay of 0.1. Models are trained for 3 epochs with a per-device batch size of 2 and gradient accumulation steps of 2, yielding an effective batch size of 32. All training is conducted in BF16 precision with gradient checkpointing enabled. The maximum sequence length is set to 65{,}536 tokens.

For DMPO, we start from \texttt{LiteCoder-Terminal}-4b-sft and train on trajectory-level preference pairs constructed from \texttt{LiteCoder-Terminal-RL}. We use DeepSpeed ZeRO-3 parallelism on 8 GPUs, with a learning rate of $5 \times 10^{-6}$, cosine learning-rate scheduling, warmup ratio 0.04, weight decay 0.1, $\beta=0.1$, $\gamma=0.7$, and 3 training epochs. The per-device training batch size is 1 with gradient accumulation steps of 4, and the maximum sequence length is 65{,}536 tokens.

\section{Broader Impacts}
\label{app:broader_impacts}

By open-sourcing \texttt{LiteCoder-Terminal-SFT}, \texttt{LiteCoder-Terminal-RL}, and \texttt{LiteCoder-Terminal}, our work lowers the barrier for building open-source terminal and software engineering agents, enabling broader participation in research and innovation. At the same time, stronger terminal agents include potential malicious or unintended uses, especially when allowed to execute commands in unconstrained environments. We therefore recommend that these models be used under human supervision and inside sandboxed execution environments with appropriate resource, network, and permission controls.



\end{document}